\pdfoutput=1

\documentclass[letterpaper, 10 pt, conference]{ieeeconf}  
\usepackage{siunitx}
\IEEEoverridecommandlockouts                              

\usepackage{graphicx}
\usepackage{amsmath} 
\usepackage{gensymb} 
\usepackage{algorithm}
\usepackage{algorithmic}
\usepackage{hyperref}
\usepackage{color}
\usepackage{siunitx}

\title{\LARGE \bf
Deep Reinforcement Learning for High Precision Assembly Tasks
}

\author{Tadanobu Inoue$^{1}$, Giovanni De Magistris$^{1}$, Asim Munawar$^{1}$, Tsuyoshi Yokoya$^{2}$ and Ryuki Tachibana$^{1}$
\thanks{$^{1}$IBM Research - Tokyo, IBM Japan, Japan. \{inouet, giovadem, asim, ryuki\}@jp.ibm.com}
\thanks{$^{2}$Tsukuba Research Laboratory, Yaskawa electric corporation, Japan. Tsuyoshi.Yokoya@yaskawa.co.jp}
}

\begin{document}

\maketitle
\thispagestyle{empty}
\pagestyle{empty}

\begin{abstract}
High precision assembly of mechanical parts requires accuracy exceeding the robot precision.
Conventional part mating methods used in the current manufacturing requires tedious tuning of numerous parameters before deployment.
We show how the robot can successfully perform a tight clearance peg-in-hole task through training a recurrent neural network with reinforcement learning.
In addition to saving the manual effort, the proposed technique also shows robustness against position and angle errors for the peg-in-hole task.
The neural network learns to take the optimal action by observing the robot sensors to estimate the system state.
The advantages of our proposed method is validated experimentally on a 7-axis articulated robot arm.
\end{abstract}

\section{INTRODUCTION}
Industrial robots are increasingly being installed in various industries to handle advanced manufacturing and high precision assembly tasks.
The classical programming method is to teach the robot to perform industrial assembly tasks by defining key positions and motions using a control box called ``teach pendant''.
This on-line programming method is usually tedious and time consuming.
Even after programming, it takes a long time to tune the parameters for deploying the robot to a new factory line due to environment variations.

Another common method is off-line programming or simulation. This method can reduce downtime of actual robots, but it may take longer time overall than on-line programming including the time for developing the simulation and testing on the robot. It is quite hard to represent the real world including environment variations with 100\% accuracy in the simulation model.
Therefore, this off-line method is not sufficient for some industrial applications such as precision machining and flexible material handling where the required precision is higher than the robot accuracy.

In this paper, we propose a skill acquisition approach where the low accuracy of conventional programming methods is compensated by a learning method without parameter tuning.
Using this approach, the robot learns a high precision fitting task using sensor feedback without explicit teaching.

\begin{figure}[t]
\centering
\includegraphics[width=0.4\textwidth]{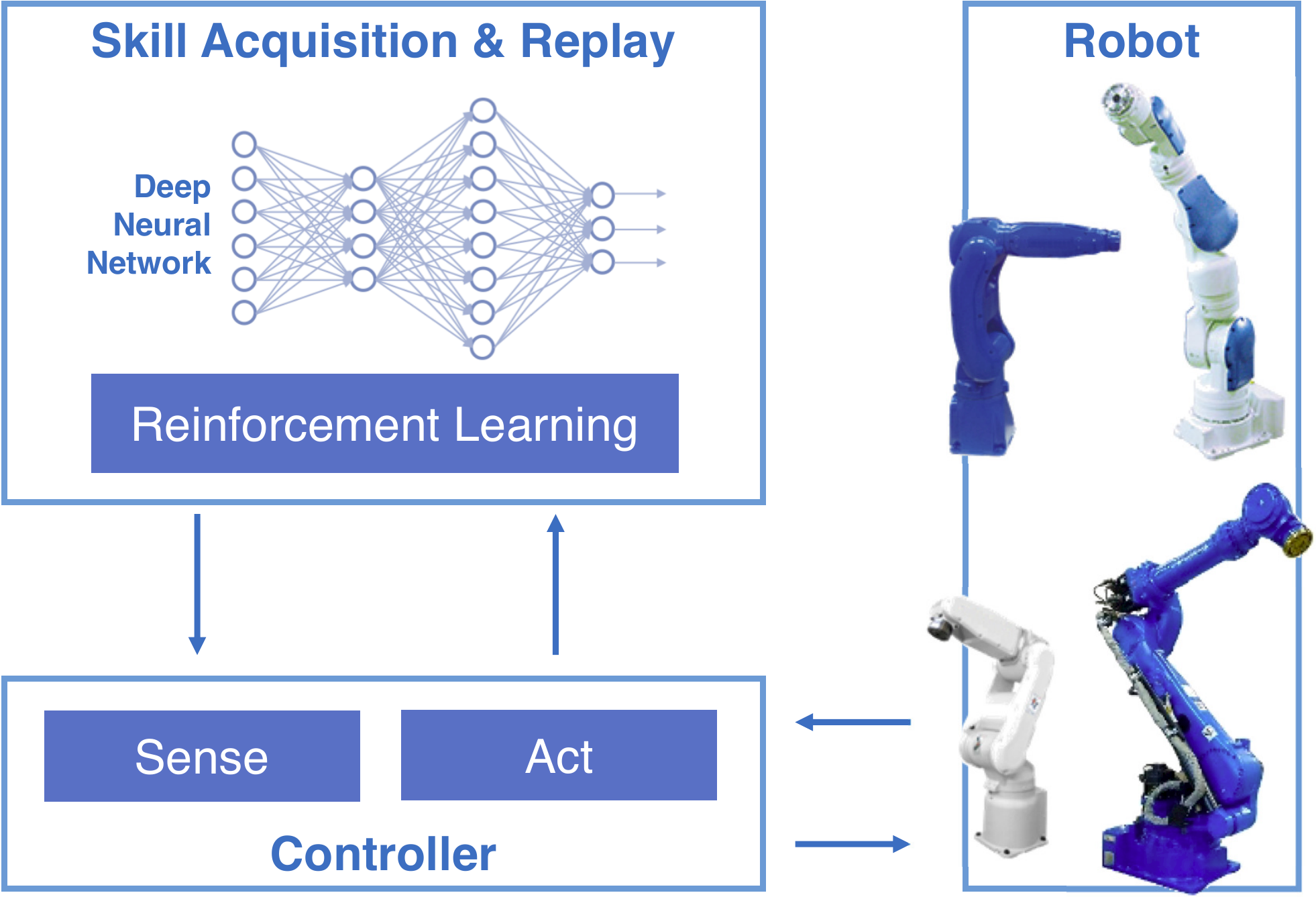}
\caption{Robot learns new skills using deep reinforcement learning}
\label{fig:concept}
\end{figure}

For such systems, reinforcement learning (RL) algorithms can be utilized to enable a robot to learn new skills through trial and error using a process that mimics the way humans learn~\cite{Kober}.
The abstract level concept is shown in Fig.~\ref{fig:concept}.
Recent studies have shown the importance of RL for robotic grasping task using cameras and encoders \cite{Levin:ISER:2016}\cite{Pinto:ICRA:2016}, but none of these methods can be applied directly to high precision industrial applications.

To show the effectiveness of this approach, we focus on learning tight clearance cylindrical peg-in-hole task.
It is a benchmark problem for the force-controlled robotic assembly.
The precision required to perform this task exceeds the robot accuracy.
In addition to tight clearance the hole can be tilted in either direction, this further adds to the problem difficulty.
Instead of using super-precise force-torque sensors or cameras, we rely on the common force and position sensors that are ubiquitous in the industrial robots.
To learn the peg-in-hole task, we use a recurrent neural network, namely, Long Short Term Memory (LSTM) trained using reinforcement learning.

The rest of the paper is organized as follows.
Section \ref{sec:problem_form} explains the problem.
Details of our proposed method is described in Section \ref{sec:rl}.
Quantitative analysis of the method on a real robot is presented in Section \ref{sec:experiments}.
Finally, we conclude the paper in Section \ref{sec:conclusions} with some directions for the future work.

\section{Problem Formulation}
\label{sec:problem_form}
A high-precision cylindrical peg-in-hole is chosen as our target task for the force-controlled robotic assembly. This task can be broadly divided into two main phases~\cite{Sharma}:
\begin{itemize}
	\item Search: the robot places the peg center within the clearance region of the hole center
	\item Insertion: the robot adjusts the orientation of the peg with respect to the hole orientation and pushes the peg to the desired position
\end{itemize}
In this paper, we study and learn these two phases separately.
\subsection{Search Phase}
Although industrial robots have reached a good level of accuracy, it is difficult to set peg and hole to few tens of \SI{}{\micro\metre} of precision by using a position controller.
Visual servoing is also impractical due to the limited resolution of cameras or internal parts that are occluded during assembly, for example, in case of meshing gears and splines in transmission.
In this paper, we use a common 6-axis force-torque sensor to learn the hole location with respect to the peg position.

Newman \textit{et al.}~\cite{Newman} calculate the moments from sensors and interprets the current position of the peg by mapping the moments onto positions.
Sharma \textit{et al.}~\cite{Sharma} utilize depth profile in addition to roll and pitch data to interpret the current position of the peg.
Although, these approaches are demonstrated to work in simulation, it is difficult to generalize them for the real world scenario.
In the real case, it is very difficult to obtain a precise model of the physical interaction between two objects and calculate the moments caused by the contact forces and friction~\cite{Bouchard:GI:2015}.

\subsection{Insertion Phase}
The insertion phase has been extensively researched. Gullapalli \textit{et al.}~\cite{Gullapalli} use associative reinforcement learning methods for learning the robot control.
Majors and Richards~\cite{Majors} use a neural network based approach.
Kim \textit{et al.}~\cite{Kim} propose the insertion algorithm which can recover from tilted mode without resetting the task to the initial state.
Tang \textit{et al.}~\cite{Tang} propose an autonomous alignment method by force and moment measurement before insertion phase based on a three-point contact model.

Compared to these previous works, we insert a peg into a hole with a very small clearance of \SI{10}{\micro\metre}.
This high precision insertion is extremely difficult even for humans.
This is due to the fact that humans cannot be so precise and the peg usually gets stuck in the very initial stage of insertion.
It is also very difficult for the robot to perform an insertion with clearance tighter than its position accuracy.
Therefore, robots need to learn in order to perform this precise insertion task using the force-torque sensor information.

\section{Reinforcement Learning with Long Short Term Memory}
\label{sec:rl}
\begin{figure}[thpb]
\centering
\includegraphics[width=0.4\textwidth]{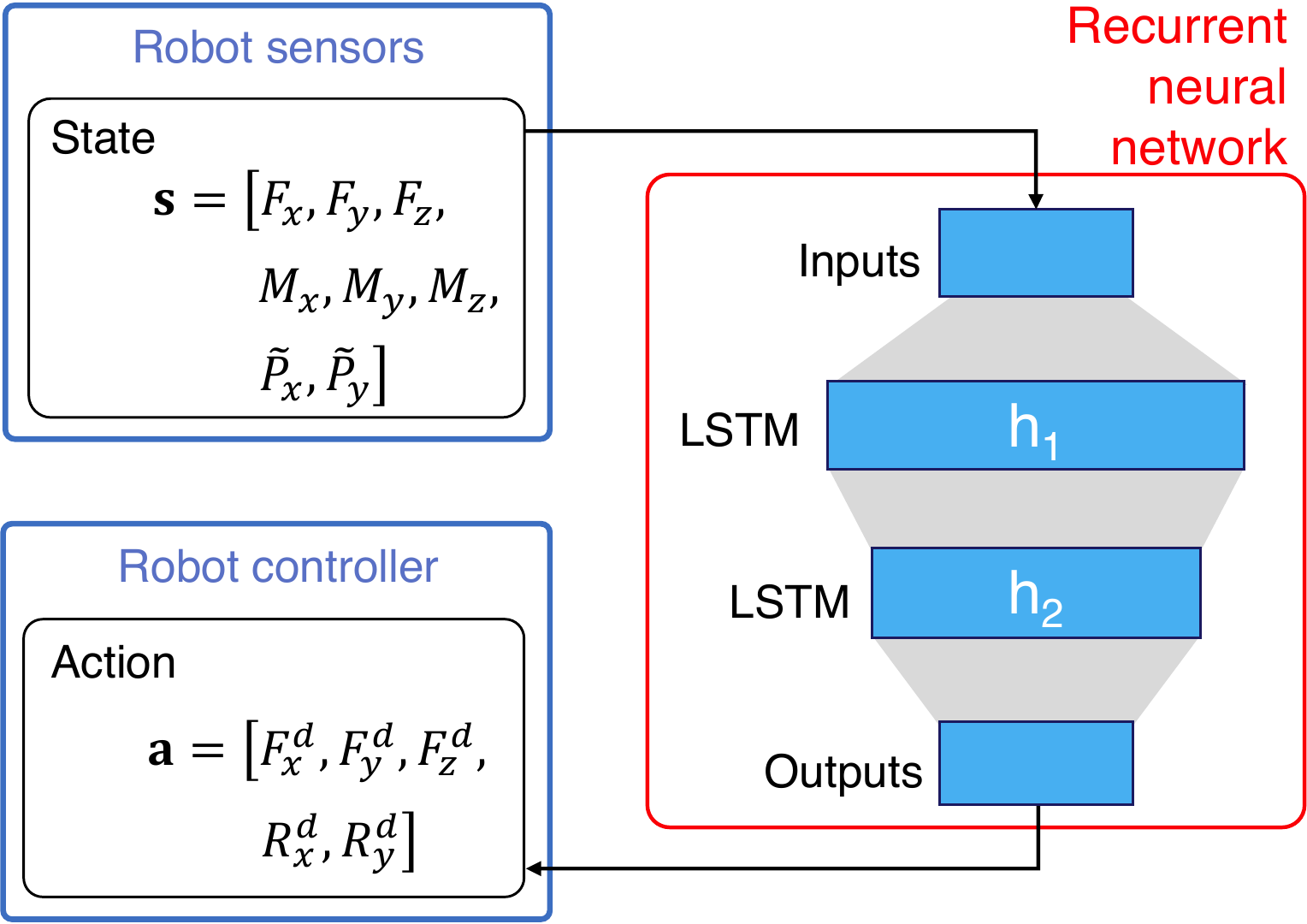}
\caption{Reinforcement learning with LTSM}
\label{fig:RL_pos}
\end{figure}

In this section, we explain the RL algorithm to learn the peg-in-hole task (Fig.~\ref{fig:RL_pos}). The RL agent observes the current state $\mathbf{s}$ of the system defined as:
\begin{equation}
\scalebox{0.95}{$
\mathbf{s} = \left[F_x, F_y, F_z, M_x, M_y, \tilde{P}_{x}, \tilde{P}_{y}\right]
$}
\label{eq:state}
\end{equation}
where $F$ and $M$ are the average force and moment obtained from the force-torque sensor;
the subscript $x,y,z$ denotes the axis.

\begin{figure}[htpb]
\centering
\includegraphics[width=0.25\textwidth]{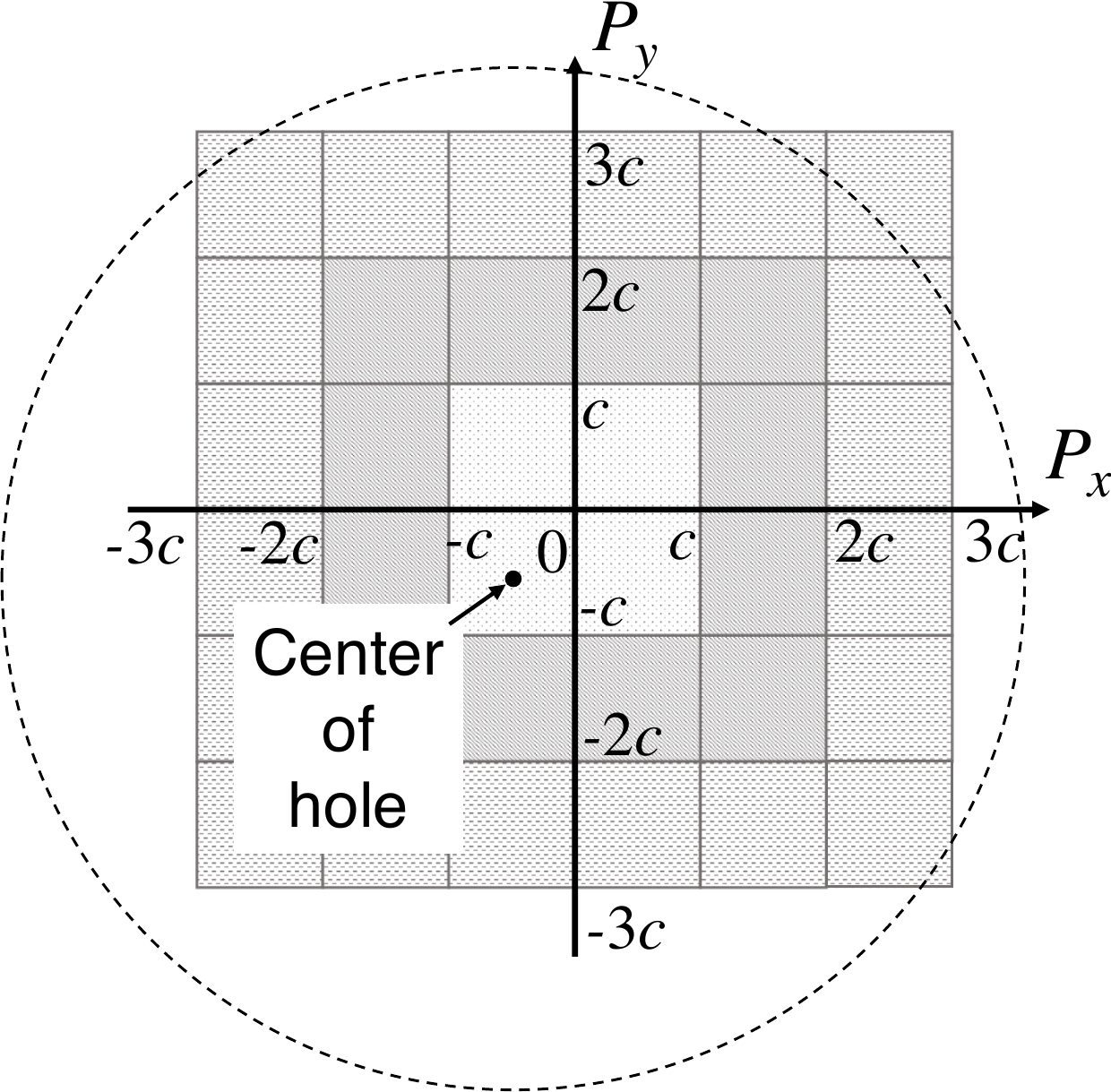}
\caption{Position data rounded to grid size}
\label{fig:data_grid}
\end{figure}

The peg position $P$ is calculated by applying forward kinematics to joint angles measured by the robot encoders.
During learning, we assume that the hole is not set to the precise position and it has position errors.
By doing this we add robustness against position errors that may occur during the inference.
To satisfy this assumption, we calculate the rounded values $\tilde{P}_x$ and $\tilde{P}_y$ of the position data $P_x$ and $P_y$ using the grid shown in Fig.~\ref{fig:data_grid}.
Instead of the origin (0, 0), the center of the hole can be located at $\mathrm{-c}<x<\mathrm{c}$, $\mathrm{-c}<y<\mathrm{c}$, where $\mathrm{c}$ is the margin for the position error.
Therefore, when the value is $(\mathrm{-c}$, $\mathrm{c})$, it will be rounded to $0$.
Similarly when the value is $[\mathrm{c}, 2\mathrm{c})$, it will be rounded to $\mathrm{c}$, and so on.
This gives auxiliary information to the network to accelerate the learning convergence.

\begin{figure}[htpb]
\centering
\includegraphics[width=0.45\textwidth]{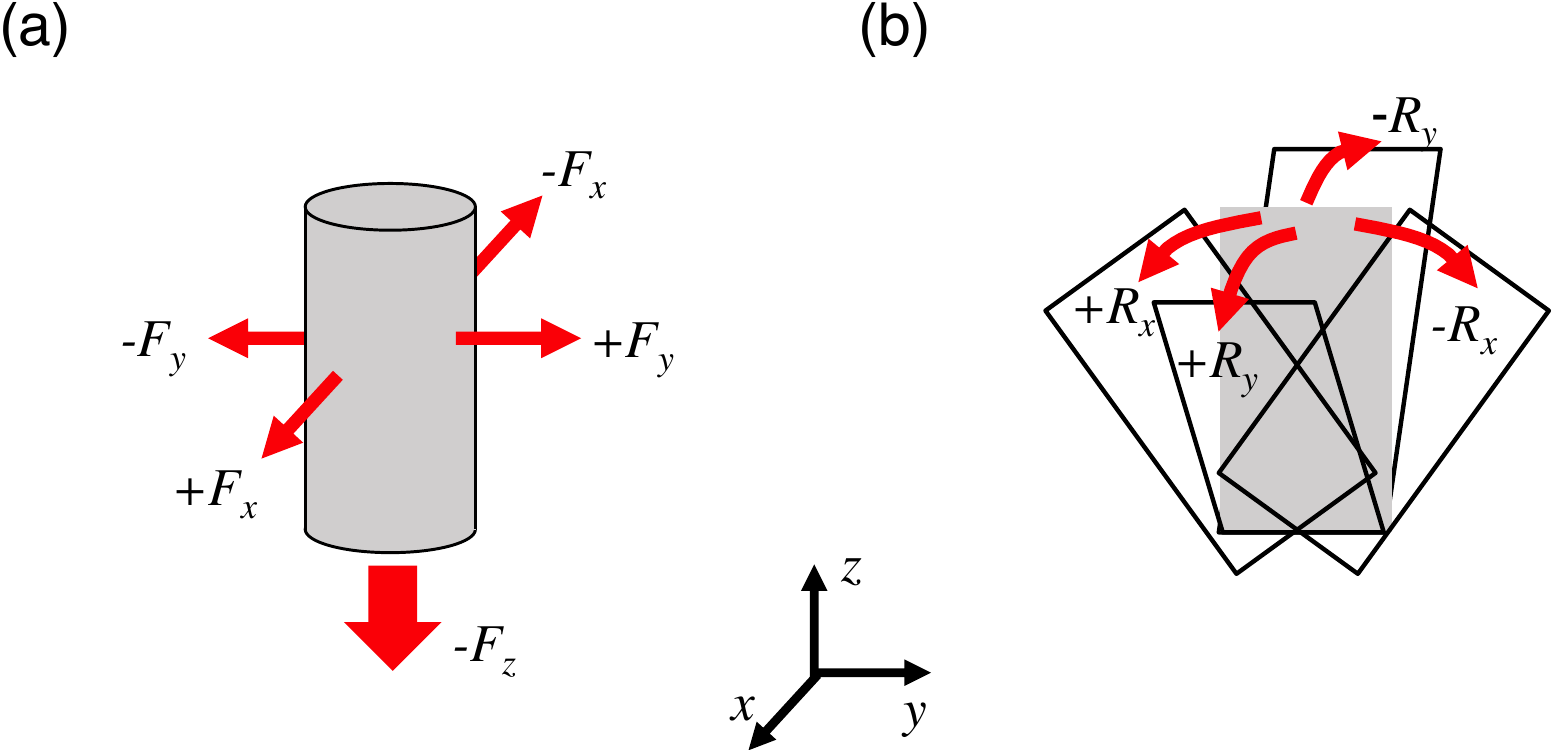}
\caption{Elementary movement: (a) Force movements (b) Rotation movements}
\label{fig:output_action}
\end{figure}

The machine learning agent generates an action $\mathbf{a}$ to the robot control defined as:
\begin{equation}
\mathbf{a} = \left[F^{d}_x,F^{d}_y,F^{d}_z,R^{d}_{x},R^{d}_{y}\right]
\label{eq:action}
\end{equation}
where, $F^{d}$ is the desired force and $R^{d}$ is the desired peg rotation given as input to the hybrid position/force controller of the manipulator.
Each component of the vector $\mathbf{a}$ is an elementary movement of the peg described in Fig.~\ref{fig:output_action}.
An action is defined as a combination of one of more elementary movements.

RL algorithm starts with a random exploration of the solution space to generate random actions $\mathbf{a}$.
By increasing exploitation and reducing exploration over time the RL algorithm strives to maximize the cumulative reward:
\begin{equation}
R_k = r_{k} + \gamma r_{k+1}  + \gamma^2 r_{k+2} + \ldots + \gamma^{n-k} r_{n} = r_{k} + \gamma R_{k+1}
\label{eq:cum_reward}
\end{equation}
where, $\gamma$ is the discount factor, $r$ is the current reward assigned to each action and $k$ is the step number.
In the proposed technique, we only compute one reward $r$ at the end of each episode.
If the trial succeeds, the following positive reward $r$ is provided to the network:

\begin{equation}
r = 1.0 - \frac{k}{\mathrm{k_{max}}}
\label{eq:positive_reward}
\end{equation}
where $\mathrm{k_{max}}$ is the maximum number of steps in one episode, $k \in \left[0, \mathrm{k_{max}}\right)$.

As we can see from Eq.~\eqref{eq:positive_reward}, the target of the learning is to successfully perform the task in minimum number of steps.
If we cannot finish the task in $\mathrm{k_{max}}$, the distance between the starting point and the final position of the peg is used to compute the penalty.
The penalty is different for search phase and insertion phase. For search phase, the penalty or negative reward is defined as:

\begin{equation}
r = \begin{cases}
0 & (d \le \mathrm{d_0}) \\
-\frac{d - \mathrm{d_0}}{\mathrm{D} - \mathrm{d_0}} & (d > \mathrm{d_0})
\end{cases}
\label{eq:negative_reward_search}
\end{equation}

where $d$ is the distance between the target and the peg location at the end of episode, $\mathrm{d_0}$ is the initial position of the peg, and $\mathrm{D}$ is the safe boundary.
For insertion phase, the penalty is defined by:

\begin{equation}
r = -\frac{\mathrm{Z} - z}{\mathrm{Z}}
\label{eq:negative_reward_insertion}
\end{equation}
where, $\rm{Z}$ is insertion goal depth and $z$ is the downward displacement from the initial peg position in the vertical direction.

The reward is designed to stay within the range of $-1 \leq r < 1$.
The maximum reward is less than $1$ because we cannot finish the task in zero steps.
The episode is interrupted with reward $-1$, if the distance of the peg position and goal position is bigger than $\mathrm{D}$ in the search phase.
In the insertion phase, the reward $r$ becomes minimum value $-1$ when the peg is stuck at the entry point of the hole.

To maximize the cumulative reward of Eq.~\eqref{eq:cum_reward}, we use a variant reinforcement learning called Q-learning algorithm.
At every state the RL agent learns to select the best possible action. This is represented by a policy $\pi(s)$:
\begin{equation}
\pi(\mathbf{s}) = \rm{argmax}_{\mathbf{a}} Q(\mathbf{s}, \mathbf{a})
\end{equation}

In the simplest case the Q-function is implemented as a table, with states as rows and actions as columns.
In Q-learning, we can approximate the table update by the Bellman equation:

\begin{equation}
  \scalebox{0.9}{$
    Q(\mathbf{s},\mathbf{a}) \leftarrow Q(\mathbf{s},\mathbf{a}) + \alpha \Bigl(r + \gamma\ \max_{\mathbf{a}'} Q(\mathbf{s}',\mathbf{a}') - Q(\mathbf{s},\mathbf{a})\Bigr)
  $}
\end{equation}
where, $\mathbf{s}'$ and $\mathbf{a}'$ are the next state and action respectively.

As the state space is too big, we train a deep recurrent neural network to approximate the Q-table.
The neural network parameters $\theta$ are updated by the following equation:

\begin{equation}
\theta \leftarrow \theta - \alpha \nabla_\theta L_\theta
\end{equation}
where, $\alpha$ is the learning rate, $\nabla$ denotes the gradient function, and $L$ is the loss function:

\begin{equation}
\scalebox{0.9}{$
\begin{array}{lll}
L_\theta &=& \frac{1}{2}\left[ \rm{target} - prediction \right]^2 \\
			 &=& \frac{1}{2}\left[ r + \gamma \max_{\mathbf{a}'} Q_\theta(\mathbf{s}',\mathbf{a}') - Q_\theta(\mathbf{s},\mathbf{a})\right]^2
\end{array}
$}
\end{equation}

Using the Q-learning update equation, the parameters update equation can be written as:

\begin{equation}
  \scalebox{0.9}{$
    \theta \leftarrow \theta + \alpha \Bigl (r + \gamma \max_{\mathbf{a}'} Q_\theta (\mathbf{s}',\mathbf{a}') - Q_\theta(\mathbf{s},\mathbf{a})\Bigr) \nabla_\theta Q_\theta(\mathbf{s},\mathbf{a})
  $}
  \label{eq:update}
\end{equation}
As shown in~\cite{Mnih_DQN}, we store the data for all previous episodes of the agent experiences to a memory pool $\mathbf{P}$ with maximum size $\mathrm{P}_{\rm{replay}}$ in a FIFO manner (Algorithm~\ref{alg:action_thread}).
Random sampling from this data provide replay events to provide diverse and decorrelated data for training.

In case of machine learning for real robot, it is difficult to collect the data and perform the learning offline.
The robot is in the loop and the reinforcement learning keep improving the performance of the robot over time.
In order to efficiently perform the data collection and learning, the proposed algorithm uses two threads, an action thread and a learning thread.
Algorithm \ref{alg:action_thread} shows the pseudo code of the action thread.
The episode ends when we successfully finish the phase, exceeds maximum number of allowed steps $\mathrm{k_{max}}$, or a safety violation occurs (i.e. going outside the safe zone $\mathrm{D}$).
It stores the observation to a replay memory and it outputs the action based on the neural network decision.
Algorithm \ref{alg:learning_thread} shows the learning thread and it updates the neural network by learning using the replay memory.

\begin{algorithm}[t]
  \caption{Action thread}
  \label{alg:action_thread}
  \begin{algorithmic}
    \STATE Initialize replay memory pool $\mathbf{P}$ to size $\mathrm{P}_{\rm{replay}}$
    \FOR{episode = 1 to M}
    \STATE Copy latest network weights $\theta$ from learning thread
    \STATE Initialize the start state to sequence $\mathbf{s}_1$
    \WHILE{NOT EpisodeEnd}
    \STATE With probability $\epsilon$ select a random action $\mathbf{a}_t$, otherwise select $\mathbf{a}_k = \rm{argmax}_\mathbf{a} Q(\mathbf{s}, \mathbf{a})$
    \STATE Execute action $\mathbf{a}_k$ by robot and observe reward $r_k$ and next state $\mathbf{s}_{k+1}$
    \STATE Store $(\mathbf{s}_k, \mathbf{a}_k, r_k, \mathbf{s}_{k+1})$ in $\mathbf{P}$
    \STATE $k = k + 1$
    \ENDWHILE
    \ENDFOR
    \STATE Send a termination signal to the learning thread
  \end{algorithmic}
\end{algorithm}

\begin{algorithm}[t]
  \caption{Learning thread}
  \label{alg:learning_thread}
  \begin{algorithmic}
    \STATE Initialize the learning network with random weights
    \REPEAT
    \IF{current episode is greater than $\mathrm{E}_{\rm{threshold}}$}
    \STATE Sample random minibatch of data $(\mathbf{s}, \mathbf{a}, r, \mathbf{s}')$ from $\mathbf{P}$. The minibatch size is $\rm{P}_{\rm{batch}}$
    \STATE Set target = $r + \gamma \max_{\mathbf{a}'} Q_\theta(\mathbf{s}',\mathbf{a}')$
    \STATE Set prediction = $Q_\theta(\mathbf{s}, \mathbf{a})$
    \STATE Update the learning network weight using equation Eq. \ref{eq:update}.
    \ENDIF
    \UNTIL{Receive a termination signal from the action thread}
  \end{algorithmic}
\end{algorithm}

Unlike~\cite{Mnih_DQN}, we use multiple long short-term memory (LSTM) layers to approximate the Q-function.
LSTM can achieve good performance for complex tasks where part of the environment's state is hidden from the agent~\cite{Bakker}.
In our task, the peg is in physical contact with the environment and the states are not clearly identified.
Furthermore, when we issue an action command shown in Eq. \eqref{eq:action}, the robot controller interprets the command and executes the action at the next cycle.
Therefore, the environment affected by the actual robot action can be observed after 2 cycles from the issuing action.
Experiments show that LSTM can compensate for this delay by considering the history of the sensed data.

\section{EXPERIMENTS}
\label{sec:experiments}

\begin{figure}[thpb]
\centering
\includegraphics[width=0.475\textwidth]{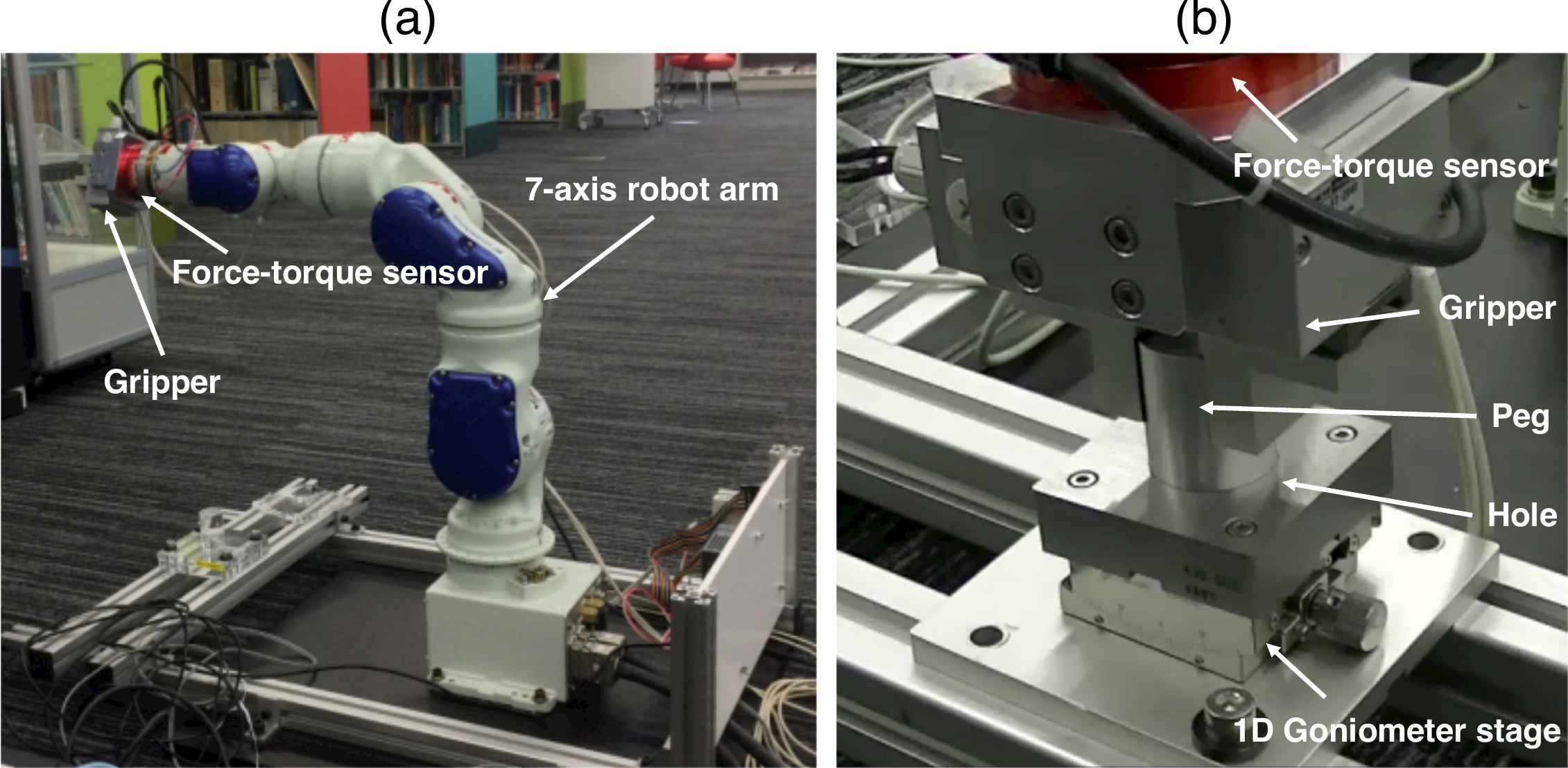}
\caption{(a) Robot (b) Description of peg-in-hole components}
\label{fig:platform}
\end{figure}

The proposed skill acquisition technique is evaluated by using a 7-axis articulated robot arm.
A 6-axis force-torque sensor and a gripper are attached to the end effector of the robot (Fig.~\ref{fig:platform}(a)).
The rated load of the force-torque sensor is \SI{200}{\newton} for the force and \SI{4}{\newton\metre} for the moment.
The resolution of the force is \SI{0.024}{\newton}.
The gripper is designed to grasp cylindrical pegs of diameter between 34 and 36 \SI{}{\milli\metre}.
In this paper, we suppose that the peg is already grasped and in contact with the hole plate.
As shown in Fig.~\ref{fig:platform}(b), a 1D goniometer stage is attached to the base plate to adjust the angle of this plate with respect to the ground.

We prepare hole and pegs with different sizes (Table \ref{tb:peg_dimension}).
The clearance between peg and the hole is defined in the table, while the robot arm accuracy is only $\pm$ \SI{60}{\micro\m}.

\begin{table}[h]
\caption{Peg and hole dimensions}
\label{tb:peg_dimension}
\begin{center}
\begin{tabular}{|c|c|c|c|c|}
\hline
Type		& Diameter 						& Height						&Material 		& Clearance\\
\hline
Peg S1 	& \SI{34.990}{\milli\metre} 	& \SI{60}{\milli\metre}  	&Steel	     	& \SI{10}{\micro\metre} \\
\hline
Peg S2 	& \SI{34.980}{\milli\metre} 	& \SI{60}{\milli\metre}  	&Steel	     	& \SI{20}{\micro\metre} \\
\hline
Hole S	& \SI{35.000}{\milli\metre} 	& \SI{20}{\milli\metre}  	&Steel			& \multicolumn{1}{r}{} 	  \\
\cline{1-4}
\end{tabular}
\end{center}
\end{table}

\begin{figure}[thpb]
\centering
\includegraphics[width=0.45\textwidth]{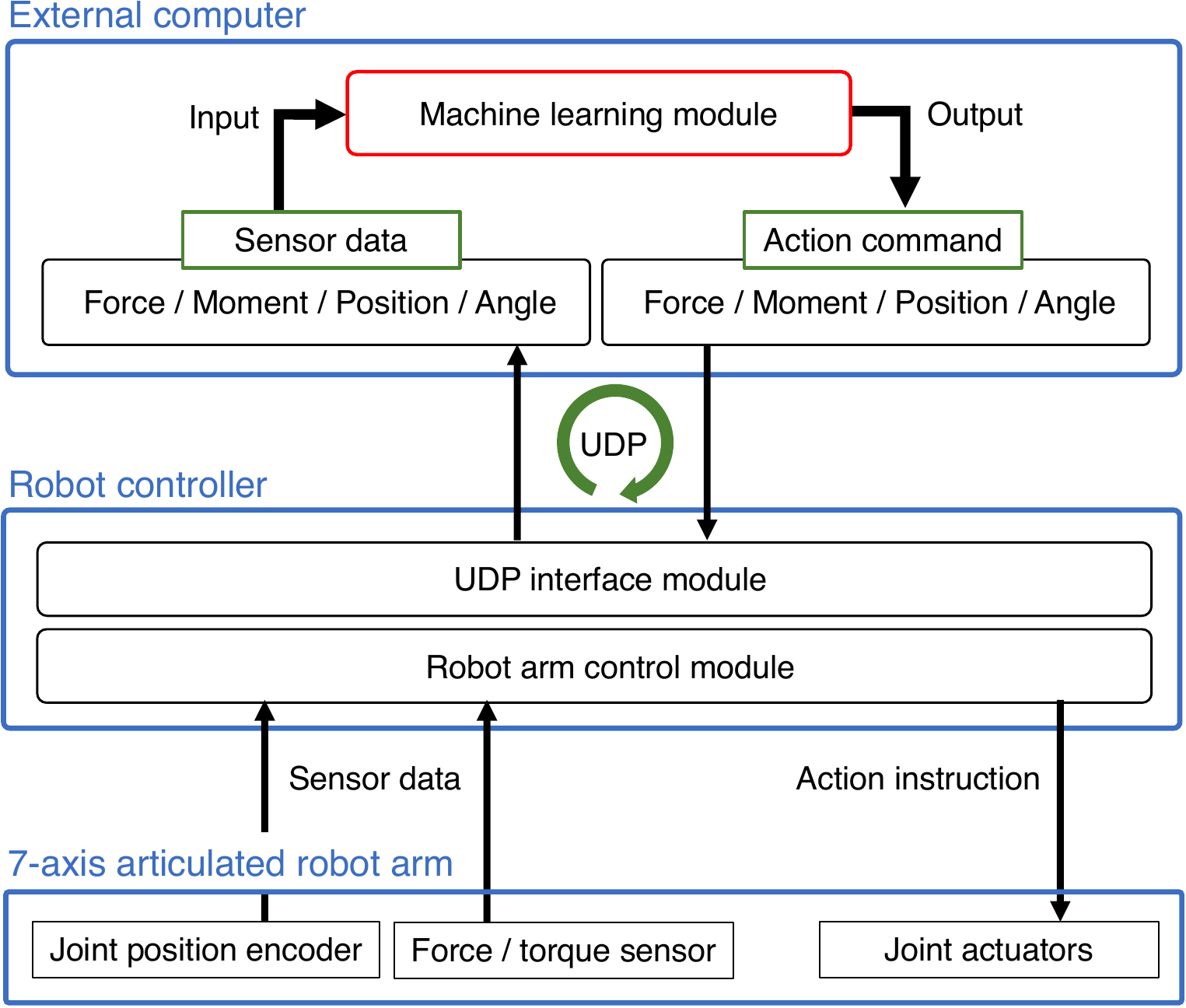}
\caption{Architecture of the experimental platform}
\label{fig:diagram}
\end{figure}

Fig.~\ref{fig:diagram} shows the architecture of the experimental platform.
The robot arm is controlled by action commands issued from an external computer (Apple MacBook Pro\textsuperscript{\textregistered}, Retina, 15-inch, Mid 2015 model with Intel Core\textsuperscript{\textregistered} i7 \SI{2.5}{\giga\hertz}).
The computer communicates with the robot controller via User Datagram Protocol (UDP).
The sensors are sampled every \SI{2}{\milli\second} and the external computer polls the robot controller every \SI{40}{\milli\second} to get 20 data points at one time.
These 20 data points are averaged to reduce the sensor noise.
The learned model is also deployed on a Raspberry Pi\textsuperscript{\textregistered} 3 for the execution.
The machine learning module in Fig.~\ref{fig:diagram} trains a LSTM network using RL to perform an optimal action for a given system state.

We treat search and insertion as two distinct skills and we train two neural networks to learn each skill.
Both networks use two LSTM layers of size $h_1 = 20$ and $h_2 = 15$ (Fig.~\ref{fig:RL_pos}).
At the first step, the search phase is learned and then the insertion phase is learned with search skill already in place.

The maximum size of the replay memory $\mathrm{P_{replay}}$ shown in Algorithm~\ref{alg:action_thread} is set to 20,000 steps and it is overwritten in a first-in-first-out (FIFO) manner.
The maximum number of episodes $M$ is set to 230 and the maximum number of steps $\mathrm{k_{max}}$ is set to 100 for the search phase and 300 for the insertion phase.
The learning thread shown in Algorithm~\ref{alg:learning_thread} starts learning after $\rm{E_{threshold}} = 10$ episodes.
Batch size is $\rm{P_{batch}} = 64$ to select random experiences from $\mathbf{P}$.

The initial exploration rate $\epsilon$ for the network is set to 1.0 (i.e. the actions are selected randomly at the start of learning).
The exploration is reduced by 0.005 after each episode until it reaches 0.1.
This allows a gradual transition from exploration to exploitation of the trained network.

\subsection{Search Phase}
\begin{figure}[t]
\centering
\includegraphics[width=0.4\textwidth]{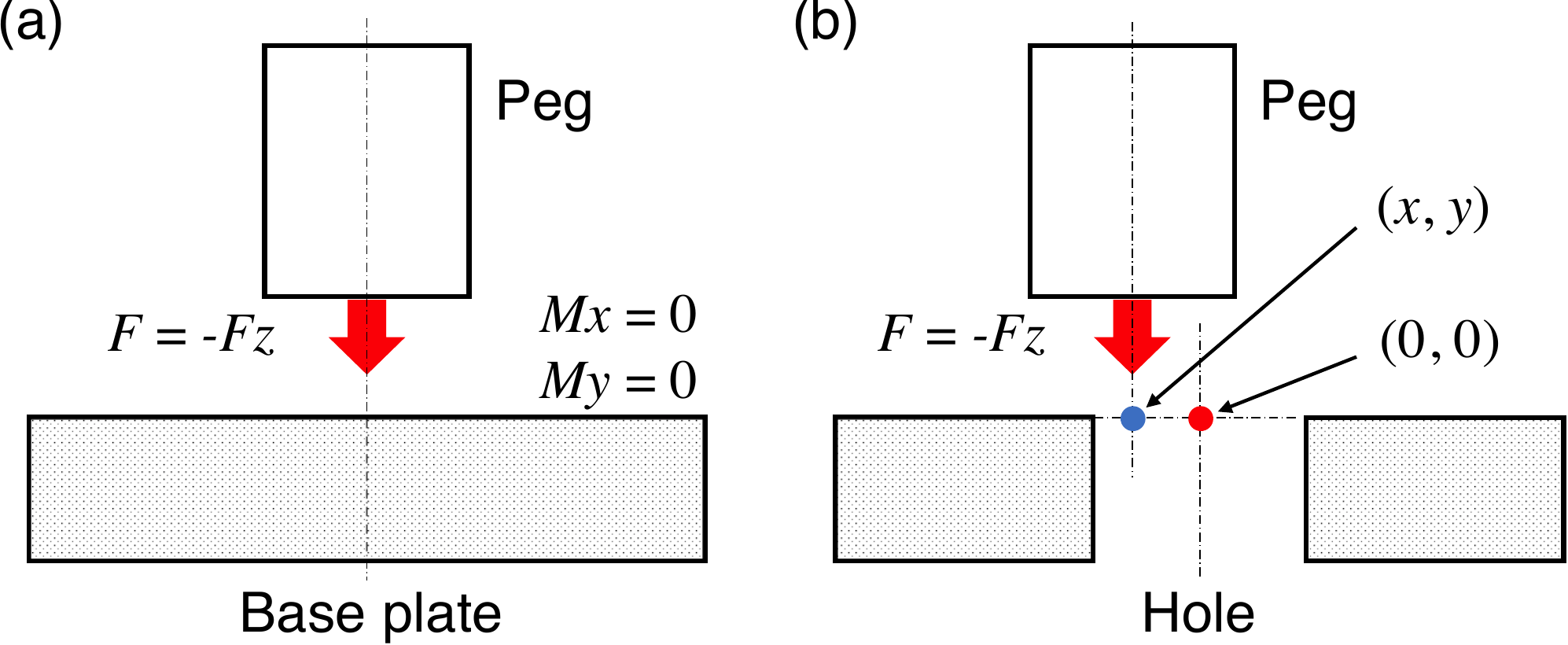}
\caption{Preliminary experiments for the moments analysis. (a) Align peg to get zero moment values (b) Stamp peg nearby the hole.}
\label{fig:preliminary}
\end{figure}

Preliminary experiments and analysis on actual robot moment were performed to compute the optimal vertical force $F^{d}_z$.
We first calibrate the 6 axis force/torque sensor.
In particular, we adjust the peg orientation ($R_x, R_y$) to ensure that both $M_x$ and $M_y$ are 0 for a vertical downward force $F_z = \SI{20}{\newton}$ (Fig.~\ref{fig:preliminary}(a)).
After calibration, we analyze the moment for three different downward forces $F^{d}_z$ at three different peg locations $(x, y)$ (Fig.~\ref{fig:preliminary}(b)).

\begin{figure}[t]
\centering
\includegraphics[width=0.48\textwidth]{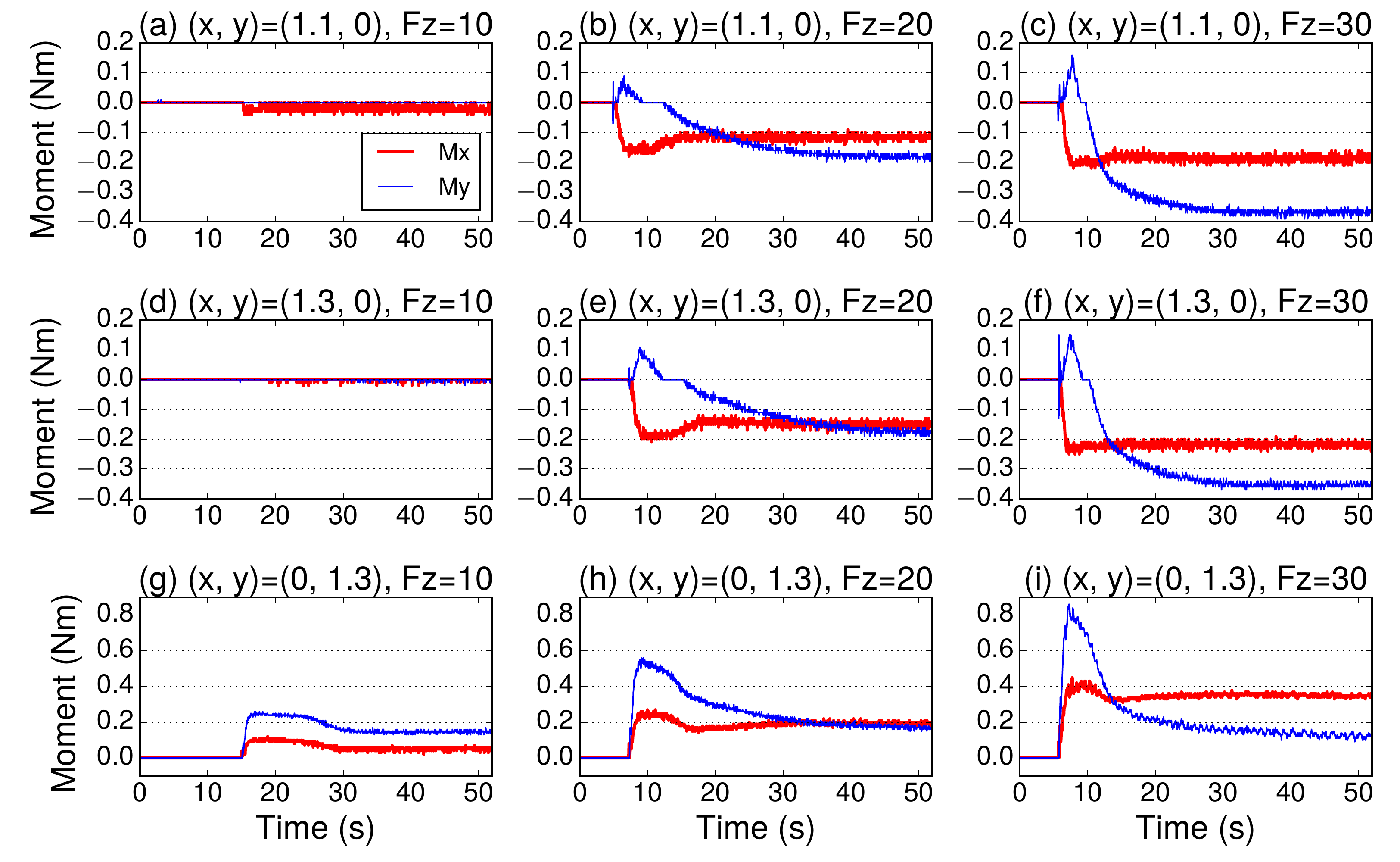}
\caption{Moment values in preliminary experiments, $M_x$ in red and $M_y$ in blue. (a)(b)(c) $(x, y)=(1.1, 0) \SI{}{\milli\metre}$ and (a) $F_z=\SI{10}{\newton}$, (b) $F_z=\SI{20}{\newton}$, (c) $F_z=\SI{30}{\newton}$; (d)(e)(f) $(x, y)=(1.3, 0) \SI{}{\milli\metre}$ and (d) $F_z=\SI{10}{\newton}$, (e) $F_z=\SI{20}{\newton}$, (f) $F_z=\SI{30}{\newton}$; (g)(h)(i) $(x, y)=(0,1.3) \SI{}{\milli\metre}$ and (g) $F_z=\SI{10}{\newton}$, (h) $F_z=\SI{20}{\newton}$, (i) $F_z=\SI{30}{\newton}$}
\label{fig:preliminary_moment}
\end{figure}

Fig.~\ref{fig:preliminary_moment} shows the moment values for nine different configurations of peg position and force.
Figs.~\ref{fig:preliminary_moment}(a) and \ref{fig:preliminary_moment}(d) show that we cannot get a detectable moment by pushing down with a force of \SI{10}{\newton}.
In contrast, it is clear that a downward force of both \SI{20}{\newton} and \SI{30}{\newton} can be used for estimating the hole direction based on the moment values.
As expected, in the case of $F^{d}_z=\SI{-20}{\newton}$ in Figs.~\ref{fig:preliminary_moment}(b) and \ref{fig:preliminary_moment}(e), $M_y$ is bigger when the peg is closer to the hole.
It is better to use a weaker force to reduce wear and tear of the apparatus, especially for relatively fragile material (e.g. aluminum, plastic).
As a result, we use \SI{20}{\newton} downward force for all subsequent experiments in search phase.

Due to the accuracy of robot sensors there is an inherent error of $\SI{60}{\micro\metre}$ in the initial position of the peg.
In addition, the hole can be set by humans manually in a factory and there can be large position errors in the initial position of the hole.
In order to make the system robust to position errors, we add additional error in the position in one of 16 directions randomly selected.
Instead of directly starting from large initial offset, the learning is done in stages for the search phase.
We start with a very small initial offset $\rm{d_0}=\SI{1}{\milli\metre}$ of the peg from the hole and learn the network parameters.
Using this as prior knowledge we increase the initial offset to $\rm{d_0}=\SI{3}{\milli\metre}$.
Instead of starting from exploration rate of 1.0 we set initial exploration rate to 0.5 for the subsequent learning stage.

The state input $\mathbf{s}$ to the search network is a 7-dimensional vector of Eq.~\eqref{eq:state}.
The size of the grid in Fig.~\ref{fig:data_grid} is set to $c = \SI{3}{\milli\metre}$ for $\rm{d_0}=\SI{1}{\milli\metre}$ and $c = \SI{5}{\milli\metre}$ for $\rm{d_0}=\SI{3}{\milli\metre}$.
The neural network selects one of the following four actions defined using Eq.~\eqref{eq:action}:
\begin{enumerate}
\item $\left[+F^{d}_x,0,-F^{d}_z,0,0\right]$
\item $\left[-F^{d}_x,0,-F^{d}_z,0,0\right]$
\item $\left[0,+F^{d}_y,-F^{d}_z,0,0\right]$
\item $\left[0,-F^{d}_y,-F^{d}_z,0,0\right]$
\end{enumerate}
with $F^{d}_x = \SI{20}{\newton}$, $F^{d}_y = \SI{20}{\newton}$ and $F^{d}_z=\SI{20}{\newton}$.
Since the peg stays in contact with the hole plate by a constant force $-F_z$, it can enter into the hole during the motion.
Compared to step wise movements, the continuous movements by the force control can avoid the static friction.

The peg position $P_z$ is used to detect when the search is successful.
If $P_z$ becomes smaller than $\rm{\Delta z_s}=\SI{0.5}{\milli\metre}$ compared to the starting point, we say that the peg is inside the hole.
We set \SI{10}{\milli\metre} for the maximum safe distance $D$ (Eq. \eqref{eq:negative_reward_search}).

\begin{figure}[thpb]
\centering
\includegraphics[width=0.48\textwidth]{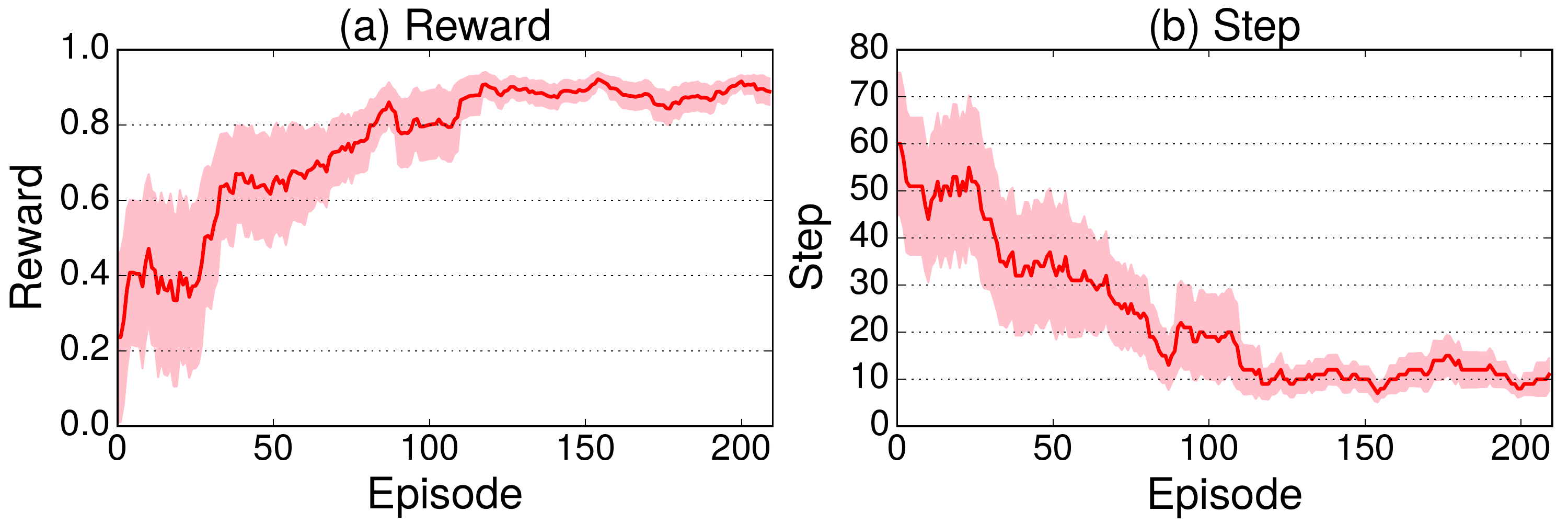}
\caption{Performance of the proposed method during learning search phase with \SI{10}{\micro\metre} clearance, \SI{0}{\degree} tilted angle, \SI{1}{\milli\metre} initial offset (a) Reward (b) Step. Means and 90\% confidence bounds in a moving window of 20 episodes}
\label{fig:learning_search}
\end{figure}

Fig.~\ref{fig:learning_search} shows the learning progress in case of \SI{10}{\micro\metre} clearance, \SI{0}{\degree} tilt angle, and \SI{1}{\milli\metre} initial offset.
Fig.~\ref{fig:learning_search}(a) shows the learning convergence and Fig.~\ref{fig:learning_search}(b) illustrates that the number of steps to successfully accomplish the search phase is reduced significantly.

\subsection{Insertion Phase}
Successful searching is a pre-requisite for the insertion phase.
After training the searching network, we train a separate but similar network for insertion.
Based on the 7-dimensional vector of Eq.~\eqref{eq:state}, we define the following state input vector of this network:

\begin{equation}
\scalebox{0.95}{$
\mathbf{s} = \left[0,0,F_z,M_x,M_y,0,0\right]
$}
\label{eq:stateinsert}
\end{equation}
where, $M_x$, $M_y$ sense the peg orientation, while $F_z$ indicates if the peg is stuck or not.

To accomplish the insertion phase, the system chooses from the following 5 actions of Eq.~\eqref{eq:action}:
\begin{enumerate}
\item $\left[0,0,-F^{d}_z,0,0\right]$
\item $\left[0,0,-F^{d}_z,+R^{d}_{x},0\right]$
\item $\left[0,0,-F^{d}_z,-R^{d}_{x},0\right]$
\item $\left[0,0,-F^{d}_z,0,+R^{d}_{y}\right]$
\item $\left[0,0,-F^{d}_z,0,-R^{d}_{y}\right]$
\end{enumerate}

The vertical peg position $P_z$ is used for the goal detection.
If the difference between starting position and the final position of the peg $P_z$ becomes larger than $\rm{Z}$, we can judge that the insertion is completed successfully.
We use \SI{19}{\milli\metre} for the stroke threshold $\rm{Z}$ (Eq. \eqref{eq:negative_reward_insertion}).
The reward for a successful episode is similar to the one used in search phase (Eq. \eqref{eq:positive_reward}).

\subsection{Results}

In order to show the robustness of the proposed technique, we perform experiments with pegs of different clearances.
We also perform tests with tilted hole plate using a 1D goniometer stage under the plate.
The results are shown in the attached video (see \url{https://youtu.be/b2pC78rBGH4}).

\begin{figure}[thpb]
\centering
\includegraphics[width=0.48\textwidth]{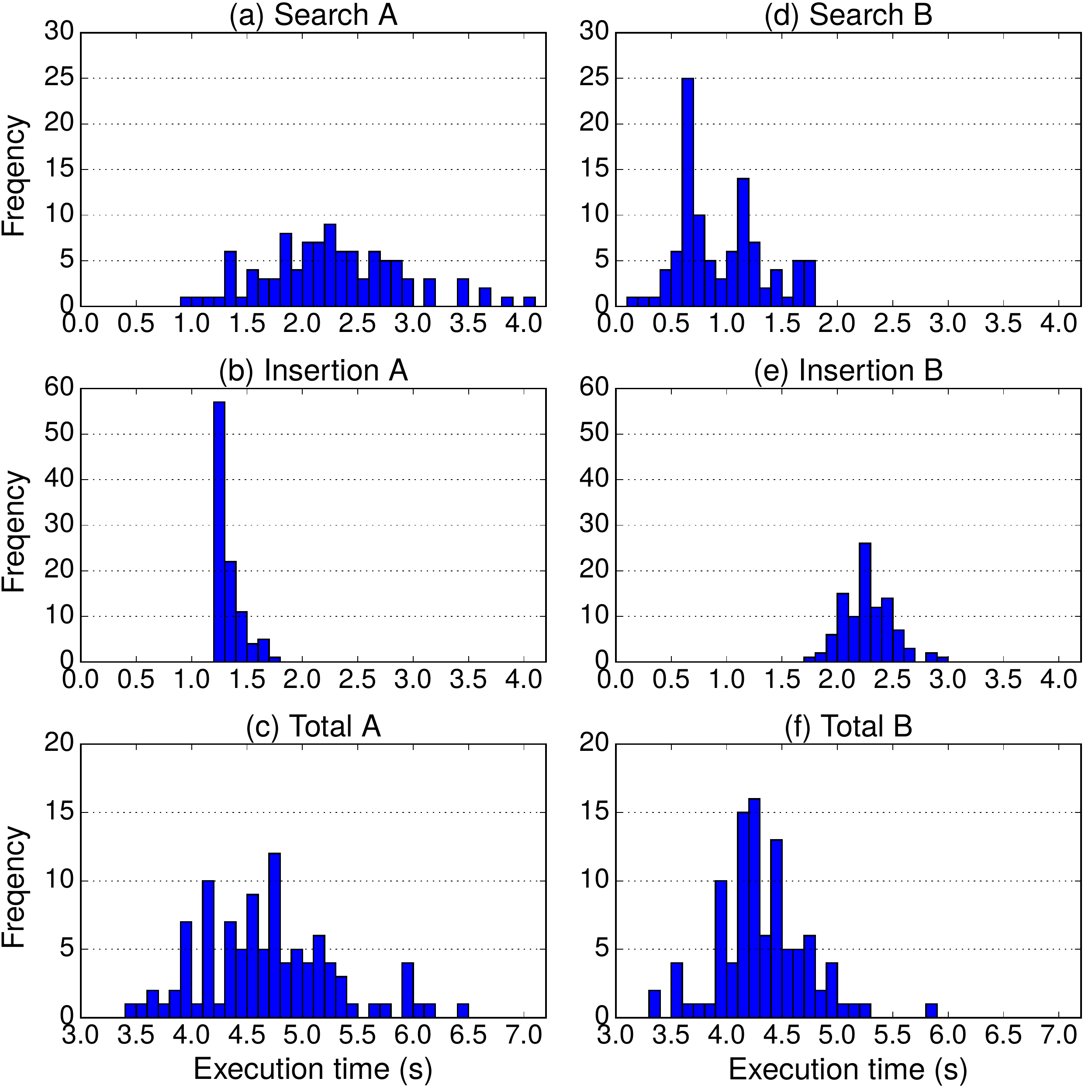}
\caption{Histograms of execution time: The case (A) that \SI{10}{\micro\metre} clearance, \SI{0}{\degree} tilted angle, \SI{3}{\milli\metre} initial offset (a) search time (b) insertion time (c) total time. The case (B) that \SI{20}{\micro\metre} clearance, \SI{1.6}{\degree} tilted angle, \SI{1}{\milli\metre} initial offset (d) search time (e) insertion time (f) total time.}
\label{fig:histogram_exec_time}
\end{figure}

We execute the peg-in-hole task 100 times after learning to show the time performances of the learning method:
\begin{itemize}
\item Case A: \SI{3}{\milli\metre} initial offset, \SI{10}{\micro\metre} clearance and \SI{0}{\degree} tilted angle
\item Case B: \SI{1}{\milli\metre} initial offset, \SI{20}{\micro\metre} clearance and \SI{1.6}{\degree} tilted angle
\end{itemize}
Fig.~\ref{fig:histogram_exec_time} shows histograms of the execution time in two cases about search, insertion, and total time.
Fig.~\ref{fig:histogram_exec_time}(a) shows the distribution of the execution time spread over wider area and is shifted further right than Fig.~\ref{fig:histogram_exec_time}(d).
When the tilt angle is larger, the execution time for the insertion becomes longer as the peg needs to be aligned with the hole.

\begin{table}[h]
\caption{Average execution time for peg-in-hole task;\newline (1) Conventional approach using fixed search patterns~\cite{MotoFit} \newline (2) Our proposed approach}
\label{tb:average_exec_time}
\begin{center}
\begin{tabular}{|c||c|c|c|c|}
\hline
Approach & (1) & (2) & (2) & (2)\\
\hline
Clearance [\SI{}{\micro\metre}] & $\geq$ 10 & 10 & 10 & 20 \\
\hline
Angle error [\SI{}{\degree}] & $\leq$ 1.0 & 0 & 0 & 1.6 \\
\hline
Initial position error [\SI{}{\milli\metre}] & $\leq$ 1.0 & 1.0 & 3.0 & 1.0 \\
\hline
Search time (s) & -- & 0.97 & 2.26 & 0.95 \\
\hline
Insertion time (s) & -- & 1.40 & 1.33 & 2.31 \\
\hline
Total time (s) & $\sim$5.0 & 3.47 & 4.68 & 4.36 \\
\hline
\end{tabular}
\end{center}
\end{table}

Table \ref{tb:average_exec_time} summarizes the average execution time in 100 trials for the four cases.
We achieve 100\% success rate in all cases.
For comparison, our results are compared with the specifications on the product catalog of the conventional approach using force sensing control and fixed search patterns~\cite{MotoFit}.
The maximum initial position and angle errors allowed by the conventional approach is \SI{1}{\milli\metre} and \SI{1}{\degree} respectively.
The results show that robust fitting skills against position and angle errors can be acquired by the proposed learning technique.

\section{CONCLUSIONS AND FUTURE WORK}
\label{sec:conclusions}

There are industrial fitting operations that require very high precision.
Classical robot programming techniques takes a long setup time to tune parameters due to the environment variations.
In this paper, we propose an easy to deploy teach-less approach for precise peg-in-hole tasks and validate its effectiveness by using a 7-axis articulated robot arm.
Results show robustness against position and angle errors for a fitting task.

In this paper, the high precision fitting task is learned for each configuration by using online learning.
In future work, we will gather trial information from multiple robots in various configurations and upload them to a Cloud server.
More general model will be learned on the Cloud by using this data pool in batches.
We would like to generalize the model so that it can handle different materials, robot manipulators, insertion angles, and also different shapes.
Then, skill as a service will be delivered to robots in new factory lines with shortened setup time.

The proposed approach uses a discrete number of actions to perform the peg-in-hole task.
As an obvious next step, we will analyze the difference between this approach and continuous space learning techniques such as A3C \cite{Mnih_A3C} and DDPG \cite{Lillicrap}.

\section*{ACKNOWLEDGMENT}

We are very grateful to Masaru Adachi in Tsukuba Research Laboratory, Yaskawa electric corporation, Japan for his helpful support to this work.

\end{document}